# TCCT: Tightly-coupled convolutional transformer on time series forecasting

Li Shen, Yangzhu Wang

*Beihang University, RM.807, 8th Dormitory, Dayuncun Residential Quarter, No.29, Zhichun Road, Beijing 100191, PR China*

A R T I C L E   I N F O

A B S T R A C T

Time series forecasting is essential for a wide range of real-world applications. Recent studies have shown the superiority of Transformer in dealing with such problems, especially long sequence time series input (LSTI) and long sequence time series forecasting (LSTF) problems. To improve the efficiency and enhance the locality of Transformer, these studies combine Transformer with CNN in varying degrees. However, their combinations are loosely-coupled and do not make full use of CNN. To address this issue, we propose the concept of tightly-coupled convolutional Transformer (TCCT) and three TCCT architectures which apply transformed CNN architectures into Transformer: (1) CSPAttention: through fusing CSPNet with self-attention mechanism, the computation cost of self-attention mechanism is reduced by 30% and the memory usage is reduced by 50% while achieving equivalent or beyond prediction accuracy. (2) Dilated causal convolution: this method is to modify the distilling operation proposed by Informer through replacing canonical convolutional layers with dilated causal convolutional layers to gain exponentially receptive field growth. (3) Passthrough mechanism: the application of passthrough mechanism to stack of self-attention blocks helps Transformer-like models get more fine-grained information with negligible extra computation costs. Our experiments on real-world datasets show that our TCCT architectures could greatly improve the performance of existing state-of-the-art Transformer models on time series forecasting with much lower computation and memory costs, including canonical Transformer, LogTrans and Informer.

## 1. Introduction

Time series forecasting plays a pivotal role in many domains, such as stock market prediction [1], event-driven sentiment analysis [2], industrial assets monitoring [3], satellite images classification [4], etc. With the arrival of the era of big data, time series forecasting models begin to face scenarios requiring longer and longer prediction length, hence for each rolling window, models should be capable of handling more past information as well. In order to meet the need of making forecasts in the long run, traditional time series forecasting models including ARIMA [5,6] and SSM [7] are not competent enough as their concrete models need manual selection to account for related various factors.

Deep neural network based model is a good candidate for solving aforementioned problems, especially Transformer models [8–11]. Compared with CNN [12–14] or RNN [15–17] time series forecasting models, Transformer's self-attention mechanism helps model equally available to any part of input regardless of temporal distance, making Transformer models more potential in dealing with long-range information and capturing long-range dependency. However, Transformer's self-attention mechanism also brings model quadratically computation cost and memory consumption growth according to the input length. It is getting worse when the model stacks several self-attention blocks. Moreover, Transformer model suffers from insensitivity to local context due to point-wise query-key matching in self-attention mechanism, making model not robust enough when receiving exception data.

For the sake of improving the efficiency and enhancing the locality of Transformer architecture, many recent researches have proposed Transformer-like models combining with CNN to solve various tasks including time series forecasting. The characteristic of local feature extraction makes CNN [18,19] complementary to Transformer. However, existing related models are mostly loosely-coupled convolutional Transformer. 'Loosely-coupled' here means that models only apply commonly used CNN layers, such as canonical convolution layers and max-pooling layers to Transformer, or otherwise sequentially or parallel stack CNN blocks and Transformer blocks. LogTrans [10] replaces linear projections of query, key and value with causal convolutional layers within

*E-mail addresses:* shenli@buaa.edu.cn (L. Shen), wangyangzhu@buaa.edu.cn (Y. Wang)

self-attention mechanism. Informer [11] uses canonical convolutional layers and max-pooling layers to connect self-attention blocks. DS-Net [20] separately generates feature maps with CNN-based network and Transformer-based network. TransCNN [21] applies pooling layers to self-attention mechanism and connects self-attention blocks with TDB and IRB blocks which are composed of depth-wise convolutional layers and max-pooling layers. TransCNN has already reached somewhere between 'loosely-coupled' method and 'tightly-coupled' method due to its usage of CNN architectures. However, its application of CNN inside the self-attention mechanism is still limited to common pooling layers.

There is no doubt that 'loosely-coupled' methods improve the performance of Transformer models to some extents. However, only 'tightly-coupled' methods, referring to applying specific transformed CNN architectures inside Transformer model, can combine Transformer and CNN tightly, thus making full use of their strengths. Thus, in this paper, we seek to answer the question: *Can specific CNN architectures be applied inside Transformer model to strengthen its learning capability and enhance its locality while improving its efficiency as well?*

To this end, our work delves into combining Transformer and CNN tightly. Three classic CNN architectures after transformation have been successfully applied to Transformer models on time series forecasting in our work. The contributions of this paper can be summarized as follows:

1. We propose the idea of tightly-coupled convolutional Transformer (TCCT) and three TCCT architectures. After transformation, these architectures not only enhance the locality of Transformer, which 'loosely-coupled' methods may also do, but also strengthen learning capability of Transformer and cut down computation cost and memory usage. They are also general enough to cope with other Transformer-like time series forecasting models.
2. We propose CSPAttention, a self-attention mechanism mirroring CSPNet belonging to CNN. It cuts down nearly 30 % of memory occupation and 50 % of time complexity of self-attention mechanism while achieving equivalent or superior forecasting accuracy.
3. We propose modified self-attention distilling operation by employing dilated causal convolution which replaces canonical convolution to connect self-attention blocks. It helps Transformer model acquire exponentially receptive field growth with a little more negligible computation cost. Therefore, the learning capability of Transformer is strengthened.
4. We propose passthrough mechanism to concatenate feature maps of different scales of self-attention blocks, thus getting more fine-grained information. Alike to feature pyramids commonly used in CNN and image processing, it expands feature maps leading to better forecasting performance of Transformer model.

## 2. Related works

### 2.1. Time series forecasting

Various methods have been proposed to solve the problem of time series forecasting due to its wide existence in many domains. Traditional time series forecasting methods are mainly based on statistics and normally have theoretical guarantees and strong interpretability [5–7,22]. There also exist methods inspired by machine learning algorithms such as support-vector machine [1] and hierarchical Bayesian method [23]. Deep learning based methods are gradually becoming the mainstream owing to the recent demand of processing and predicting long sequence multivariate time series which traditional methods are unable to undertake. Prevalent deep learning based methods are mainly based on RNN [17,24–26]. They achieve better performances compared with traditional models especially dealing with multivariate time series forecasting problems, but still not good enough when facing long sequence time series input (LSTI) or long sequence time series forecasting (LSTF) problems. In order to seek and establish the long-range dependency between outputs and inputs, thus solving LSTI and LSTF problems, Transformer is a good choice thanks to its self-attention mechanism.

### 2.2. Transformer models on time series forecasting

Several Transformer models have been proposed in recent years to solve time series forecasting problems. They are mostly derived from Vaswani Transformer [27] and make little changes essentially [9,28,29]. LogTrans and Informer are two works mostly related to our work due to their effort on combination of CNN and Transformer. LogTrans replaces linear projections of self-attention mechanism with convolutional layers and Informer uses convolutional layers and max-pooling layers to connect self-attention blocks, meaning that they are both loosely-coupled convolutional Transformer models. This is their main limitation which our work is intended to address. However, it cannot be denied that they are state-of-the-art, therefore, our experiments are mainly based on these two baselines to examine the effect of our three TCCT architectures and upgrade them to tightly-coupled convolutional Transformer models, especially the more competitive Informer.

### 2.3. Related CNN

Convolutional neural networks [18,30–32] hold a pivotal position when tackling with computer vision problems. Among so many great works, we would like to highlight two CNN architectures that are deeply related to our work, CSPNet [33] and Yolo series networks [34–36]. Although computer vision and time series forecasting are completely different tasks, some great ideas from these two CNN architectures can be borrowed. Cross Stage Partial Network (CSPNet) intends to mitigate the computation bottleneck of complex CNN architectures to handle heavy computer vision tasks. CSPNet alleviates the problem of duplicate gradient information by separating feature maps at the beginning of the stage, then making parts of feature maps directly linked to the end of the stage and integrating all of them at last. Similar concept is applied to self-attention mechanism in our paper. Yolo series convolutional networks are very famous real-time object detectors. Even now, state-of-the-art Yolo-based or Yolo-related object detectors are continuously being proposed, such as PP-YOLOv2 [37], YOLOS [38], etc. Instead of applying whole Yolo baseline into Transformer architecture, we only borrow the idea of passthrough mechanism firstly proposed in Yolo9000 [34] and apply it to the stack of self-attention blocks within the encoder of Transformer. Apart from computer vision, CNN also holds a foothold in time series forecasting tasks, such as Wavenet [13], TCN [12], Seq-U-net [14], etc. The core idea of these networks is the usage of causal convolutional layer and its offspring forms. In our work, we borrow the concept of dilated causal convolution from Temporal Convolutional Network (TCN) to connect self-attention blocks in order to gain exponentially receptive field growth and enhance the locality of Transformer.

## 3. Preliminary

### 3.1. Problem definition

Before introducing Methodology, we first provide the definition of time series forecasting problem. Suppose we have a fixed input window $\{z_{i,1:t_0}\}_{i=1}^{N}$, the task is to predict corresponding fixed target window $\{z_{i,t_0+1:t_0+T}\}_{i=1}^{N}$. N refers to the number of related univariate time series, t0 is the input window size and T is the prediction window size. Given a long target sequence whose length is much longer than preset output window size, rolling forecasting strategy is adopted to predict the whole sequence.

### 3.2. Informer architecture

Informer is a time series forecasting model derived from canonical Transformer [27]. Several improvements are made by Informer to enhance the prediction capability of Transformer-like models in long sequence time series forecasting (LSTF) problem. Due to its state-of-the-art performance, we utilize it as the main baseline and apply our TCCT architectures to it. We would briefly introduce its major changes from canonical Transformer and refer readers to [11] for more details. Four points are concluded.

A. Instead of canonical self-attention mechanism, the encoder of Informer contains ProbSparse self-attention mechanism. ProbSparse self-attention allows each key only attend to dominant queries while performing the scaled dot-product to improve the efficiency. The way to judge the dominant queries is through the Kullback-Leibler divergence of query-key attention probability distribution and the uniform distribution. Queries owning larger KL divergence are regarded as more dominant ones. Due to the long tail distribution of self-attention scores, ProbSparse self-attention only needs to calculate $O(\ln L_Q)$ dot-products instead of $O(L_Q)$ in canonical self-attention.

B. More than above, self-attention distilling operation is performed within Informer's encoder. Convolutional layers and max-pooling layers are used to connect self-attention blocks and downsample the output of the previous self-attention block into its half slice. In this way, while stacking $k$ self-attention blocks, the length of the final output of the encoder would be $1/2^{k-1}$ of the input length.

C. To enhance the robustness, Informer could alternatively contain $k$-1 extra encoders with progressively decreasing input length and number of self-attention blocks. The $i^{th}$ extra encoder stacks $k$-$i$ self-attention blocks and the input of it is the last $1/2^i$ of the input of the main encoder so that the output lengths of all encoders are the same. We call this alternative method full distilling operation for following usage.

D. Except for changes above, Informer also contain a generative-style inference decoder predicting outputs by one forward procedure. The input vector of the decoder is composed of two parts, start token and target sequence. The masked self-attention mechanism of decoder is replaced by masked ProbSparse self-attention mechanism.

The total architecture of Informer with full distilling operation and 3-layer main encoder is shown in Fig. 1.

## 4. Methodology

We first sequentially introduce three tightly-coupled convolutional Transformer (TCCT) architectures: CSPAttention, dilated causal convolution and passthrough mechanism. Then the method to combine TCCT architectures with Transformer-like time series forecasting models would be shown.

### 4.1. CSPAttention

The architecture of one block of our proposed CSPAttention is shown in Fig. 2. The input $\in R^{L\times d}$, where $L$ is the input length and $d$ is the input dimension, is split into two parts through dimension $X = \left[X_1^{L\times d_1}, X_2^{L\times d_2}\right]$. $X_1$ is linked to the end of the block after going through a 1 × 1 convolutional layer A while $X_2$ plays the role of the input of the self-attention block B. The outputs of A and B are concatenated through dimension as the output of the whole block.

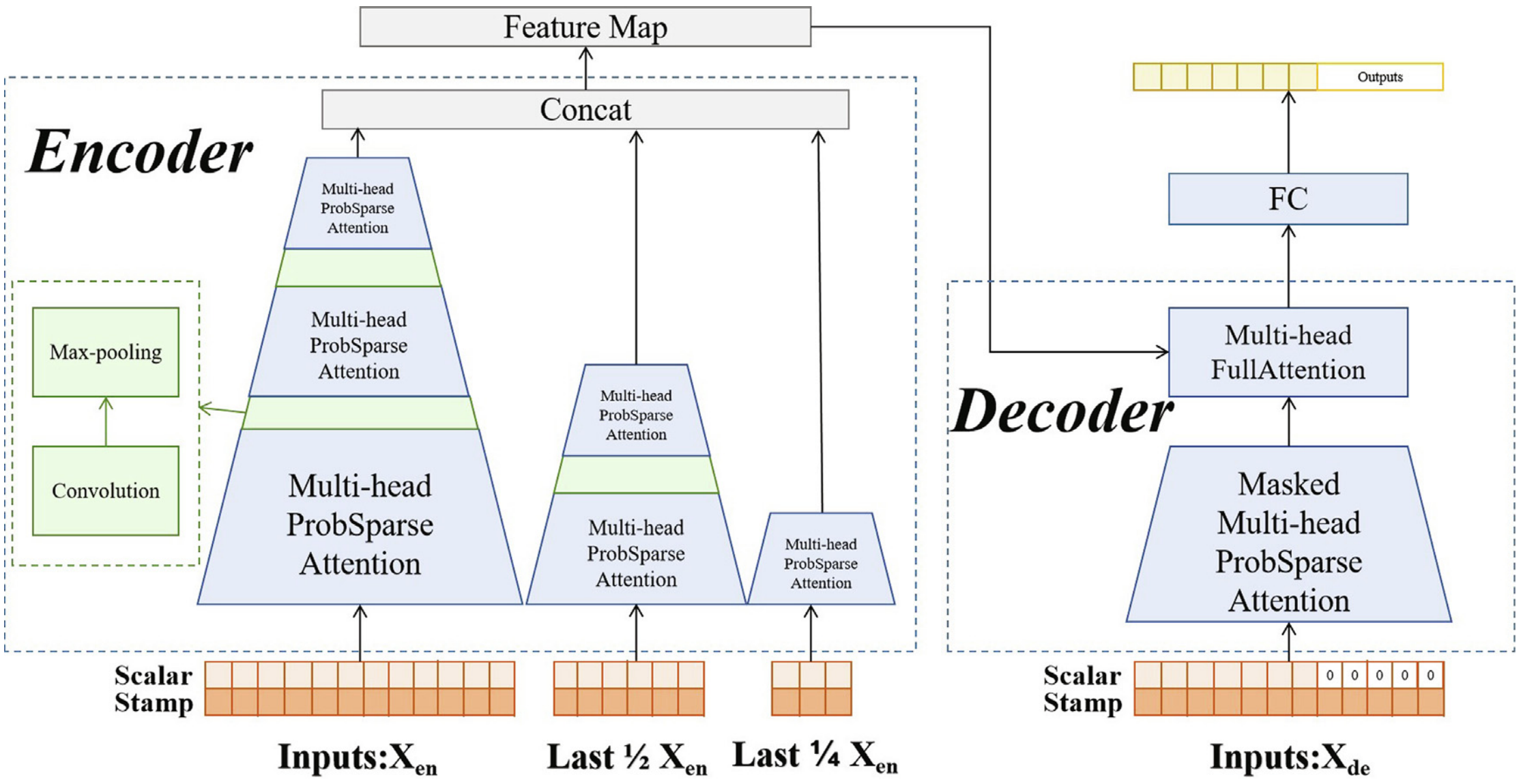

**Fig. 1.** An example of Informer architecture with full distilling operation. Left: Inside the blue trapezoid includes a main encoder stacking three ProbSparse Attention blocks (Blue) and two extra encoders with shorter inputs and fewer attention blocks. A convolutional layer (Green) together with a max-pooling layer inside the green trapezoid is used to connect each two self-attention blocks. All three feature maps outputted by three encoders are fused and then given to the decoder. Right: The decoder of Informer in the blue trapezoid almost remains unchanged compared with canonical Transformer's decoder. Only the masked self-attention block is replaced by masked ProbSparse Attention block. (For interpretation of the references to colour in this figure legend, the reader is referred to the web version of this article.)

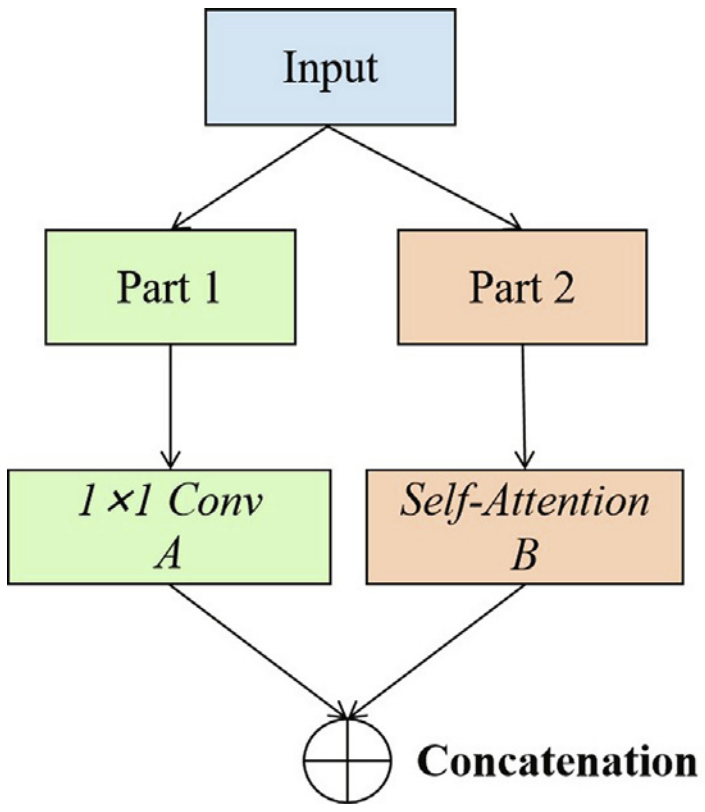

**Fig. 2.** A CSPAttention block. The input (Blue) is split into two parts. The first one (Green) propagates through layer A, a 1 × 1 convolutional layer, while the other (Red) propagates through block B, a self-attention block. The outputs of both parts are concatenated together at last as the final output of the whole CSPAttention block. (For interpretation of the references to colour in this figure legend, the reader is referred to the web version of this article.)

Excluding biases, the output matrix of one-stage of CSPAttention is given by

$$Y = \begin{bmatrix} Y_1 \\ Y_2 \end{bmatrix} = \begin{bmatrix} X_1 W_c \\ Concat(A(X_{2h})W_h)_{h=1}^{H} \end{bmatrix}$$

where (1) $A(X_{2h})$ is the scaled dot-product of the $h^{th}$ self-attention block. $W_h$ is a $d_h \times d_h$ linear projection matrix. (2) $H$ is the number of heads and $d_h$ is the dimension of each head supposing that each head has the same dimension. (3) $W_c$ is a $d/2 \times d/2$ value weight matrix of $1 \times 1$ convolutional layer.

Inspired by CSPNet [33], the way CSPAttention treats the dimensions of the tokens mirrors how CSPNet treats the channels of the images. The whole output $Y$ could be written as a sub-block matrix so that the split two parts do not contain duplicate gradient information that belongs to the other. Moreover, in order to project the first part to the proper dimension when the whole output dimension is not equal to the whole input dimension, we add an additional $1 \times 1$ convolutional layer. The weight updating of CSPAttention is shown as below:

$$W_h' = f(g_2, W_h)$$

$$W_c' = f(g_1, W_c)$$

where $f$ is the weight updating function and $g$ represents the gradient propagated to the $i^{th}$ path. It can be seen that the gradients of split parts are separately integrated.

The purpose of designing CSPAttention is to mitigate the problems of memory bottleneck and computing efficiency of self-attention mechanism. CSPNet has already shown its capability in enhancing the performance and reducing the computations of CNN based architectures. Our CSPAttention also reduces memory traffic and time complexity of self-attention mechanism. Assuming that the input and output dimension of a canonical self-attention block are both $d$ and there is only one input token. As is shown in Fig. 3(a), a self-attention block contains four linear projection layers whose input and output dimensions are both $d$ (Query, Key, Value, Projection). Therefore, the memory occupation is $4d^2$. However, supposing that CSPAttention splits the input dimensions in half, the first part of CSPAttention has only one linear projection layer, while the second has four. The corresponding architecture is shown in Fig. 3(b). Thus, the memory occupation of a CSPAttention block is $(4+1) \times (d/2)^2$, which is 31.25 % of that of a canonical self-attention block.

**Theorem1.** *Supposing that CSPAttention splits the input dimension in half, it cuts down at least* 50 % *of time complexity compared with a canonical self-attention block.*

The proof is deferred to Appendix A. Theorem1 implies that CSPAttention not only reduces the memory occupation, but also cuts down sizable time complexity.

CSPAttention can also cope with other Transformer-like architectures and upgrade them to tightly-coupled convolutional Transformer architecture. We take LogTrans [10] as an example, the combined architecture——LogSparse CSPAttention block is shown in Fig. 4. LogTrans's optimization centers on the length of query during the scaled dot-product, however, CSPAttention's optimization revolves around the dimension of the input. They are independent of each other so that in Fig. 4 LogSparse self-attention can directly replace the canonical self-attention in Fig. 2. Similarly, when applied CSPAttention to Informer, the ProbSparse self-attention would be upgraded to ProbSparse CSPAttention. More importantly, similar to CSPNet, CSPAttention increases the number of gradient paths ensuring model could achieve equivalent or superior forecasting accuracy with less computation and memory usage.

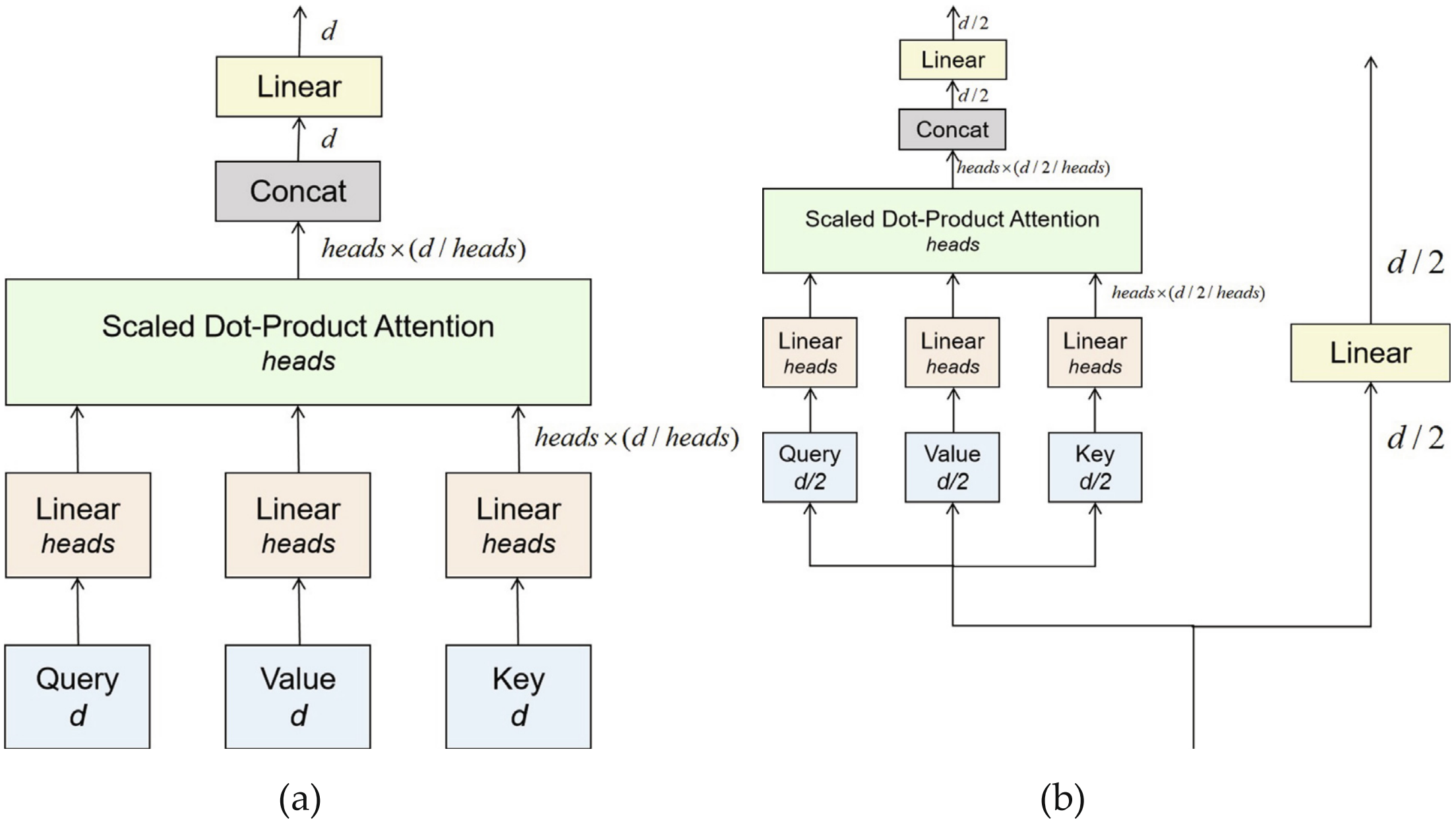

**Fig. 3.** Comparison of canonical self-attention block and CSPAttention block. The left (a) is a classic multi-head self-attention architecture assuming that each head has the same dimension. The right (b) a CSPAttention block that splits the input by dimension in half. The first left part is actually a self-attention block with half the input dimension, the other right part is linked to the end of the whole block through a 1 × 1 convolutional layer.

### 4.2. Dilated causal convolution

Stacking multiple self-attention blocks is beneficial to extract deeper feature maps, however, it brings out more time and space complexity. In order to further reduce the memory usage, Informer puts self-attention distilling operation into use. Informer employs a convolutional layer and a max-pooling layer between each two self-attention blocks to trim the input length. Convolutional layer of kernel size 3 with stride 1 follows the former self-attention block to make features more aware of local context information. Max-pooling layer of kernel 3 with stride 2 is then used to privilege the local dominating features and give a lesser but more focused feature map to the latter self-attention block. Graphically, a network stacking three self-attention blocks is shown in Fig. 5. We only show the receptive field of the last input element for simplicity.

However, canonical convolutional layer has two dominant drawbacks when applied to time series forecasting. First, it is only capable of looking back linear-size-history as the depth of the network grows. Therefore, even Informer is intended to deal with long sequence time series forecasting problems, it is not strong enough in handling extremely long sequences. Stacking self-attention blocks together with canonical convolutional layers will not bring enough benefits with the growth of computing cost, and in contrary it may cause repetitive and meaningless computation because of limited receptive field. In addition, canonical convolutional layer does not put temporal perspective into consideration which will inevitably lead to future information leakage in time series forecasting.

Our solution, inspired from TCN [12], replaces canonical convolution with dilated causal convolution to acquire exponentially receptive field growth. More formally, for the $i^{th}$ convolutional layer following the $i^{th}$ self-attention block, the dilated causal convolution operation $C$ of kernel size $k$ on element $x_n \in R^d, n \in L$ of the sequence $X \in R^{L\times d}$ is defined as

$$C(x_n) = \begin{bmatrix} x_n \\ x_{n-i} \\ \cdot \\ \cdot \\ \cdot \\ x_{n-(k-1)\times i} \end{bmatrix} W^{d\times d'}$$

$d'$ is the output dimension and the number $i$ here also works as dilation factor. The $i^{th}$ dilated causal convolutional layer's filter skips ($2^{i-1}$-1) elements between two adjacent filter taps. Moreover, each element $x$ at time $t$ of the sequence is only convolved with elements at or before $t$ due to the nature of causality, ensuring that there would be no information leakage from the future. Notice that dilated causal convolution is degraded to normal causal convolution when $i$ = 1. We provide an illustration of a network stacking three self-attention blocks and employing dilated causal convolutional layers of kernel 3 in Fig. 6.

Comparing Figs. 5 and 6, it can be clearly observed that dilated causal convolution only uses padding at the temporally front side, preventing the leakage of future information. Even only with two convolutional layers, the output receptive field of the network in Fig. 6 is noticeably larger than that of Fig. 5. Consequently, with more self-attention blocks stacked, the gap would be more enormous and hence, the disparity of two networks' performances would be greater. Beyond that, the application of dilated causal convolution only brings a little more computing cost and memory usage (padding cost), which could be ignored.

When coupled with CSPAttention, dilated causal convolutional layer also works as a transition layer attending to fuse feature maps from two parts of the former CSPAttention block.

### 4.3. Passthrough mechanism

Feature pyramid is commonly used to extract features in computer vision and CNN [40]. A similar concept could be applied to transformer based network. The passthrough mechanism, proposed by Yolo series object detection CNN networks [34–36], takes feature maps from earlier network and merges them with final feature map to get more finer-grained information.

We employ it to merge different scales of feature maps in Transformer based network. Suppose that an encoder stacks $n$ self-attention blocks, then each self-attention block would produce a feature map. The $k^{th}$ ($k$ = 1,2…$n$) feature map has the length of $L/2^{k-1}$ and the dimension of $d$ assuming that CSPAttention and dilated causal convolution have already been applied to this encoder. In order to concatenate all different scales of feature maps, the $k^{th}$ feature map is split equivalently by length into $2^{n-k}$ number of feature maps with length $L/2^{n-1}$. In this way, all feature maps could be concatenated by dimension. However, the concatenated feature map would has dimension of $(2^n - 1) \times d$, so a transition layer should be adopted to ensure the whole network exports feature map of proper dimension. We present a network stacking three self-attention blocks and employing all TCCT architectures mentioned above as Fig. 7.

The function of passthrough mechanism is similar to full distilling operation of Informer. However, Informer with full distilling operation needs encoders as many as the number of self-attention blocks of the main encoder while Informer with passthrough mechanism only needs a single encoder. Even though the encoders of Informer with full distilling operation have progressively decreasing input lengths, it gives rise to model's more attention to later time sequence data. For instance, suppose that Informer stacks $k$ encoders, then the first half of the input sequence only exists in the main encoder, oppositely, the last $1/2^{k-1}$ of the input sequence exists in every single encoder. Passthrough mechanism does not have deficiency like this. More importantly, passthrough mechanism almost brings Informer no extra computation cost while the Informer with full distilling operation would engender considerable extra computation cost because of its multi-encoder architecture.

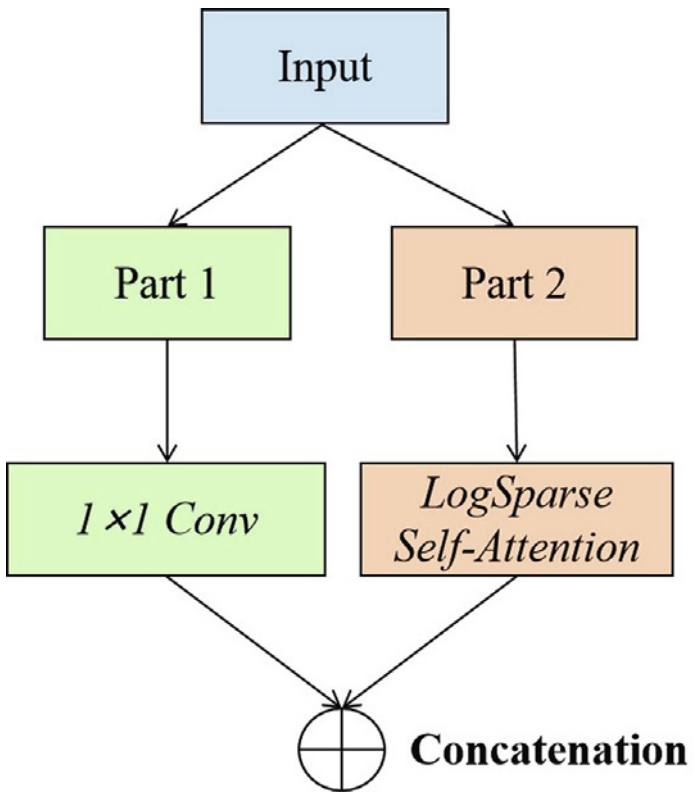

**Fig. 4.** A LogSparse CSPAttention block. Compared with Fig. 2, the canoncial self-attention block (Red) is replaced with LogSparse self-attention block. (For interpretation of the references to colour in this figure legend, the reader is referred to the web version of this article.)

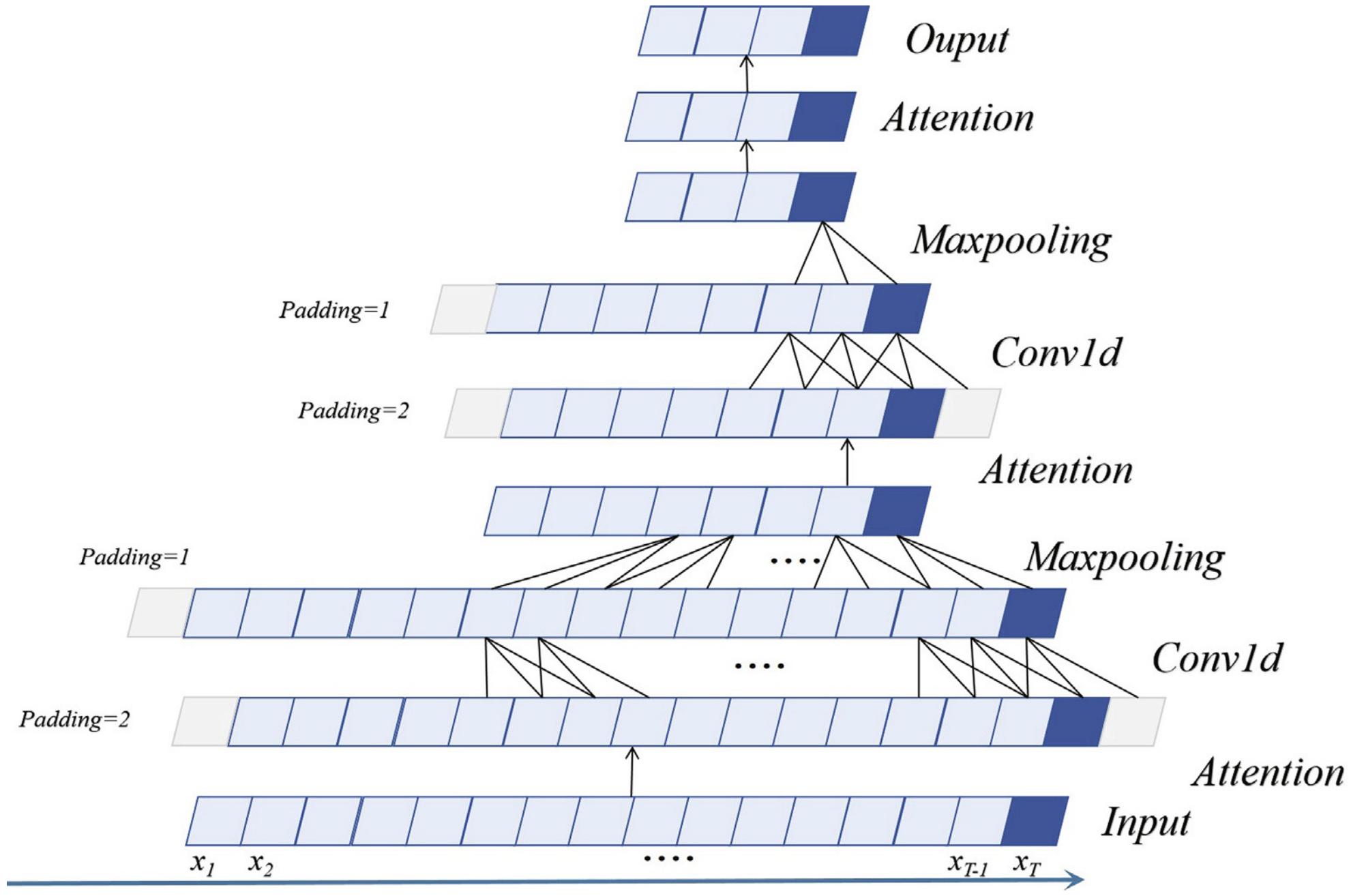

**Fig. 5.** A visualization of a self-attention network stacking three self-attention blocks connected with convolutional layers and max-pooling layers. Each layer's receptive field is reflected by the elements (Light Blue) having connections with the former layer represented by black lines. We only show the receptive field of the last element (Dark Blue) of the output as an example. (For interpretation of the references to colour in this figure legend, the reader is referred to the web version of this article.)

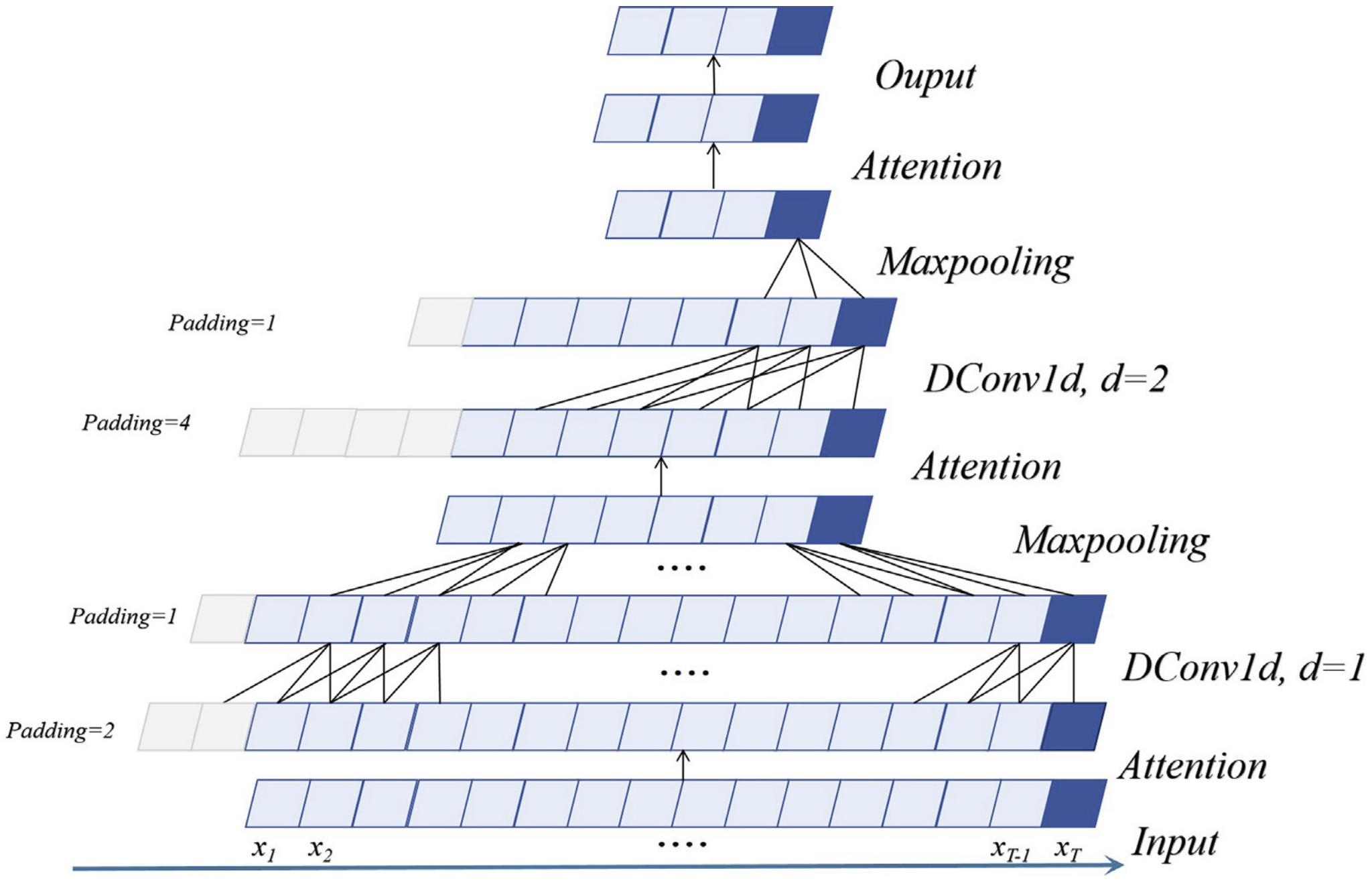

**Fig. 6.** A visualization of a self-attention network stacking three self-attention blocks connected with dilated causal convolutional layers and max-pooling layers. The whole architecture is similar to that in Fig. 5. However, the application of dilated causal convolutional layers gives rise to wider receptive field and the avoidance of future information leakage.

### 4.4. Transformer with TCCT architectures

All above architectures can seamlessly cooperate with Transformer or Transformer-like time series forecasting models, including canonical Transformer, LogTrans, Informer, etc. A simple example of cooperation with Informer is shown in Fig. 8 and a detailed encoder example is shown in Fig. 9. Note that the Informer in Fig. 8 only has one encoder, meaning that it does not use the full distilling operation and replaces it with passthrough mechanism.

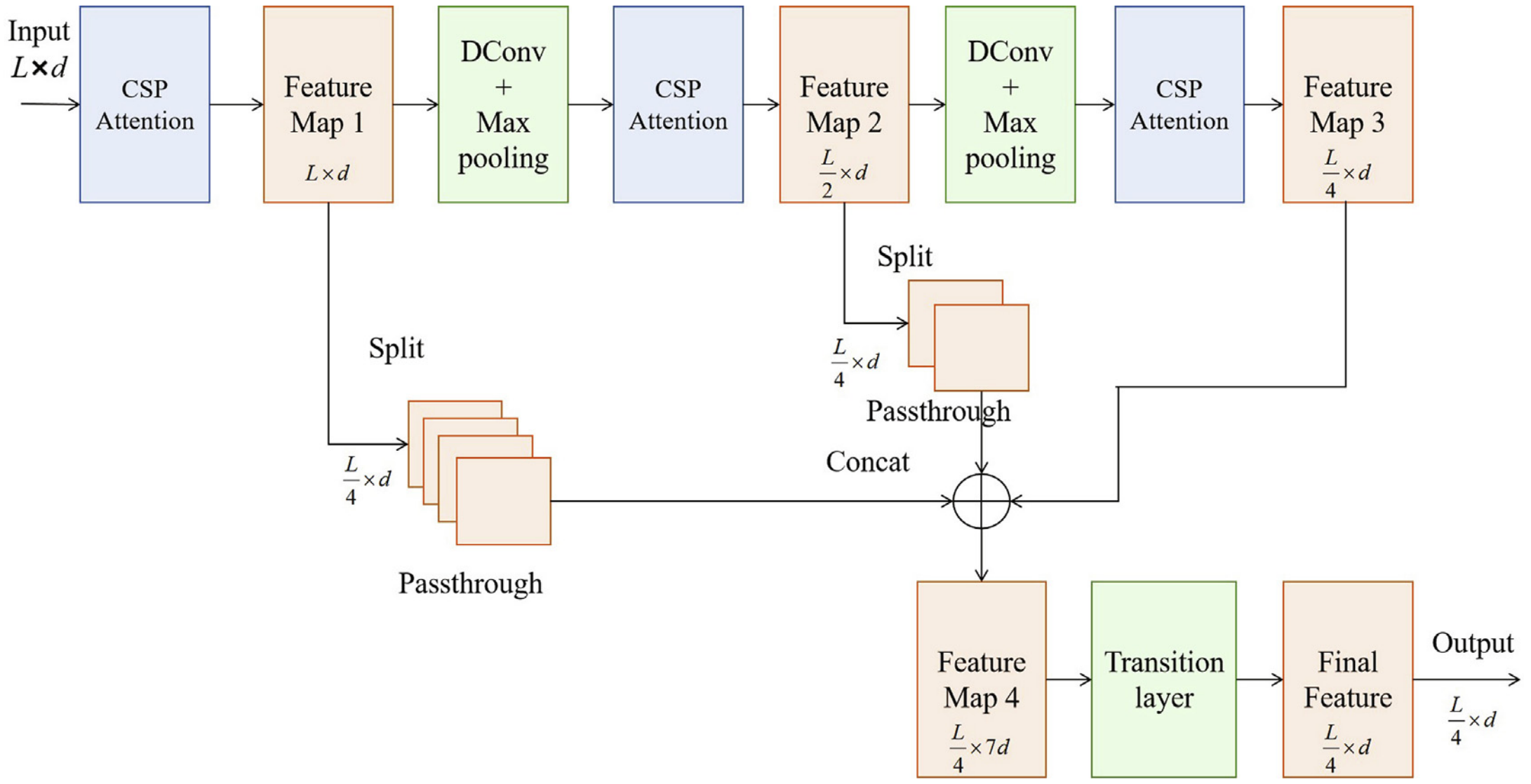

**Fig. 7.** A network stacking three CSPAttention (Blue) blocks. Dilated causal convolution (Green above) and passthrough mechanism are employed. The final output has the same dimension with input's. (For interpretation of the references to colour in this figure legend, the reader is referred to the web version of this article.)

On the basis of Informer architecture, other Transformer-like baselines could easily cooperate with TCCT architectures. For instance, to combine TCCT architectures with LogTrans, just (masked) ProbSparse self-attention blocks in Fig. 8 will be replaced by (masked) LogSparse self-attention blocks and other architectures remain the same.

## 5. Experiment

### 5.1. Datasets

We mainly perform experiments on a public real-world **ETT**[1] (Electricity Transformer Temperature) dataset composed of ETT data lasting for almost 2 years. ETT dataset consists of four subsets: {ETTh1, ETTh2} are 1-hour-level datasets from two separated counties in China; {ETTm1, ETTm2} are 15-min-level datasets from same sources as {ETTh1, ETTh2}. Each data point is composed of the target value 'oil temperature' and other 6 power load features. Raw data with Z-score normalization are used during network training/validation/testing. We mainly choose ETTh1 and ETTm1 to evaluate models. The train/val/test is 12/4/4 months as default.

More specifically, the datasets above are split evenly, consecutively and compactly across time. The training subset contains the first 12-month data, the validation subset contains the next 4-month data and the testing subset contains the last 4-month data. More detailed information of datasets is shown in Appendix D.

### 5.2. Methods

We choose Informer as the basic baseline and respectively test the effect of our proposed TCCT architectures coping with it. Informer has shown its state-of-the-art performance by experiments while compared with many other excellent works on time series forecasting, including ARIMA [41], Prophet [42], LSTMa [43], etc. So it is quite meaningful to extensively research how much our TCCT architectures could improve Informer. Therefore, five methods are chosen: Informer, the basic Informer with only one encoder; Informer+, Informer with full distilling operation; TCCT_I, Informer combined with CSPAttention; TCCT_II, Informer combined with CSPAttention and dilated casual convolution; TCCT_III, Informer combined with all TCCT architectures. Without exceptional instructions, Informer, TCCT_I, TCCT_ II, TCCT_ III contain an encoder stacking 3 self-attention blocks, while Informer + contains a 3-encoder stack with full distilling operation. In addition, every method contains a 2-layer decoder. Among these five methods, TCCT_I and TCCT_ II are mainly compared with Informer for their one-encoder architecture. TCCT_ II is mainly compared with Informer + in virtue of their multi-encoder architecture or passthrough mechanism which has similar function. Moreover, to further study the applicability of our proposed TCCT architectures to enhancement of other Transformer or Transformer-like models on time-series forecasting, we select two additional methods, canonical Transformer and LogTrans, and study whether our TCCT architectures could enhance their performances when coping with them. The source code is available at https://github.com/OrigamiSL/TCCT2021.[2]

### 5.3. Experiment details

Four groups of experiments are conducted to inspect the improvement on forecasting accuracy and efficiency TCCT architectures bring to Transformer or Transformer-like models on time-series forecasting. MSE is chosen as the loss function. All methods are optimized with Adam optimizer with learning rate starting from 1e to 4 and decaying two times smaller every epoch. The start token length of the decoder keeps identical to the input length of the encoder. The total number of epochs is 6 with proper early stopping. Batch size is set as 32. Each single experiment is individually repeated for ten times and average results are taken. All experiments are conducted on a single Nvidia GTX 1080Ti 12 GB GPU. Further concrete details are shown in concrete experiments.

### 5.4. Result and analysis

#### 5.4.1. Ablation study on LSTF problem

Under this setting, the time series forecasting capability of five methods are evaluated on both univariate and multivariate condi-tions so that how much forecasting accuracy-three TCCT architectures could respectively enhance Informer will be illustrated. ETTh1 and ETTm1 datasets are both used for examination. We fix the encoder input length as 384 and prolong the prediction window size progressively in {48, 96, 192, 384}.MSE $= \frac{1}{n}\sum_{i=1}^{n}(y-\hat{y})^2$ and MAE $= \frac{1}{n}\sum_{i=1}^{n}|y-\hat{y}|$(averaging for multivariate prediction) are chosen to compare the forecasting accuracy of different methods. $n$ refers to the prediction window size. These two criteria are also chosen to evaluate the forecasting accuracy in the following experiments if needed. Due to the average of ten-turn experiment results, we specially give and analyze MSE results distribution and consistency through corresponding standard deviation ($S = \sqrt{\frac{\sum(y-\hat{y})^2}{n}}$) and coefficient of variation ($CV = \frac{S}{\hat{y}}$) in this group of experiments. The results and analysis are shown in Appendix B. The analysis of MAE or another groups' results could be neglected in that they show similar characteristics.

[1] ETT dataset was acquired at https://github.com/zhouhaoyi/ETDataset

[2] These codes are complementary to initial Informer (https://github.com/zhouhaoyi/Informer2020). Use them together.

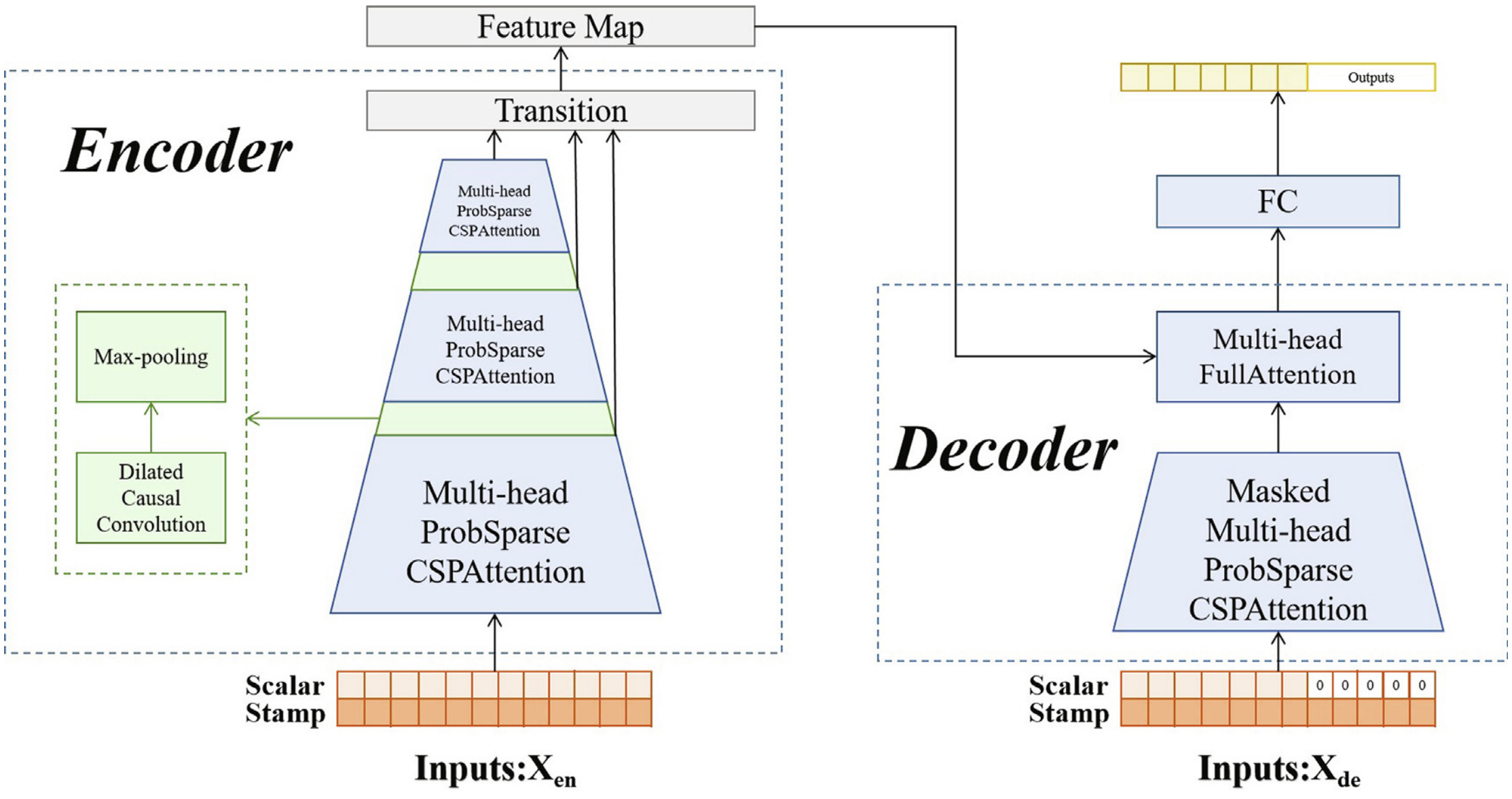

**Fig. 8.** Overview of Informer combining with proposed TCCT architectures. Left: Inside the blue trapezoid is the encoder stacking three ProbSparse CSPAttention blocks (Blue) which replace the previous ProbSparse self-attention blocks in Fig. 1. A dilated causal convolutional layer (Green), rather than a canonical convolutional layer, together with a max-pooling layer inside the green trapezoid is used to connect each two self-attention blocks. No extra encoders are added and all three feature maps outputted by three self-attention blocks are fused and then transited to the final output of proper dimension. Right: No distinguishing changes are made compared with the decoder of Informer in Fig. 1. Only the masked ProbSparse self-attention block is replaced by masked ProbSparse CSPAttention block. (For interpretation of the references to colour in this figure legend, the reader is referred to the web version of this article.)

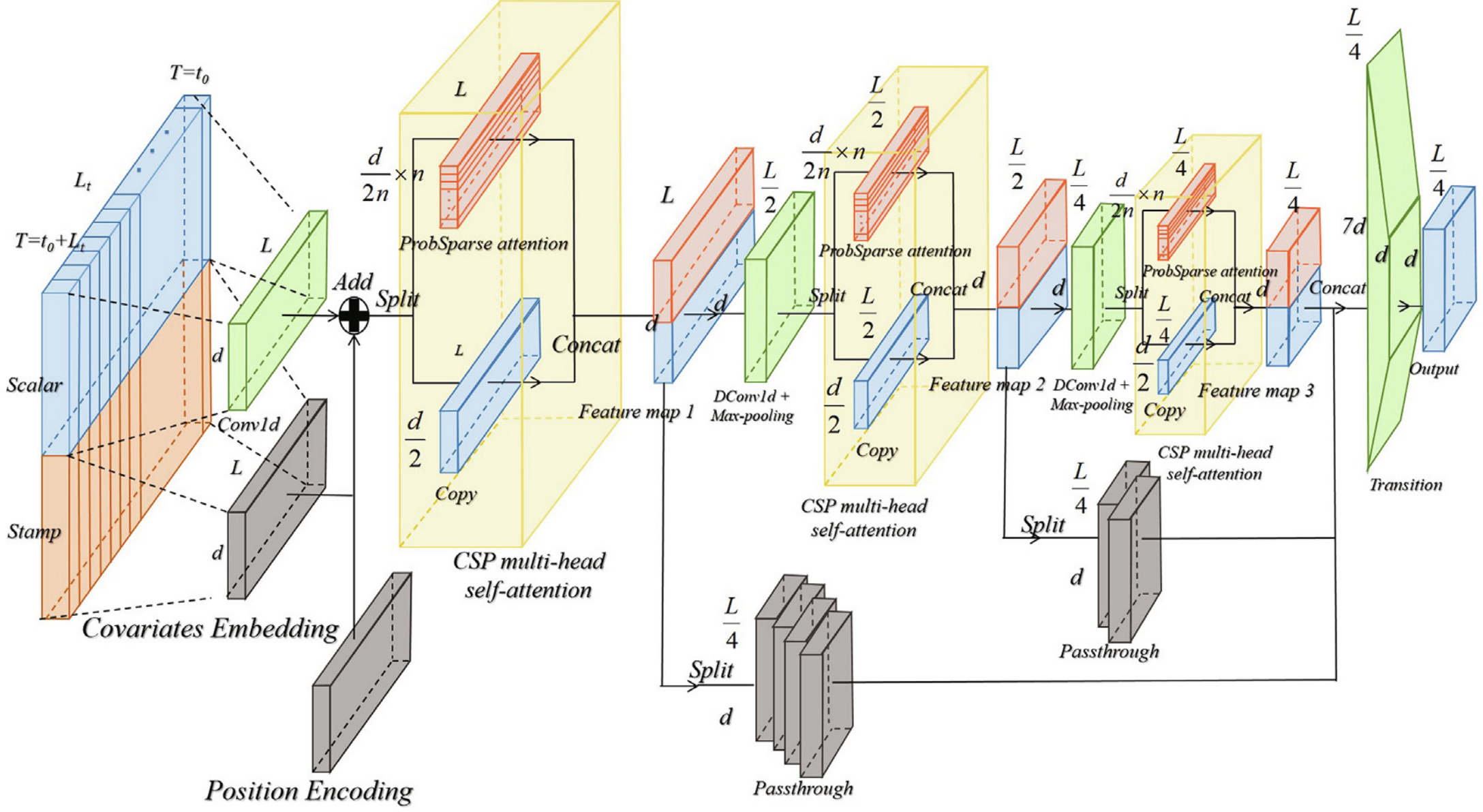

**Fig. 9.** A single Informer's encoder stacking three self-attention blocks cooperated with all three TCCT architectures. (1) Each CSPAttention block (Light Yellow) is combined with ProbSparse self-attention (Red) which is Informer's typical architecture. (2) Between every-two CSPAttention blocks, a dilated causal convolutional layer and a max-pooling layer (Green) are used for connection. The former self-attention block's output feature map shrinks to half its length after propagating through these two layers, mirroring the situation in initial Informer but enlarging the receptive field. (3) All three feature maps outputted by three self-attention blocks are fused by passthrough mechanism (Gray) to acquire finer-grained information. A transition layer (Green) is added in the end to export feature map of proper dimension to the decoder. (For interpretation of the references to colour in this figure legend, the reader is referred to the web version of this article.)

From Tables 1 and 2, we can observe that: (1) TCCT_ III, Informer with all three TCCT architectures, outperforms Informer+, Informer with full distilling operation, at most of cases under univariate setting and all cases under multivariate setting, demonstrating that our proposed three architectures truly enhance the prediction capability of Informer in LSTF problem. (2) Compared with Informer, TCCT_I, Informer with CSPAttention, shows equivalent performances at few cases and superior performances at most cases under both univariate and multivariate setting, demonstrating that CSPAttention could help Informer gain more lightweight architecture without losing forecasting accuracy. (3) TCCT_ II, Informer with both CSPAttention and dilated causal convolution, outperforms Informer and TCCT_I at nearly all cases, illustrating that the application of dilated causal convolutional layers further improves the prediction capability on the basis of TCCT_I. When it comes to multivariate conditions, TCCT_ II even outperforms Informer + at nearly half cases, especially when the prediction lengths are 192 and 384. (4) TCCT_ III outperforms all other four methods, demonstrating the benefit of the application of passthrough mechanism to Informer. It also shows that passthrough mechanism is more effective and more robust than full distilling operation. (5) Under univariate setting, Informer + outperforms TCCT_I and TCCT_ II, proving that the method of adding encoders works. It even has 25 % chance of outperforming TCCT_ III under the current experiment environment. However, when it comes to multivariate conditions, Informer + is totally defeated by TCCT_ III and starts to lose to TCCT_ II as the prediction length grows. This phenomenon demonstrates that as the complexity of forecasted sequence grows, our proposed TCCT architectures become stronger in enhancing the prediction capability of Informer than full distilling operation. Therefore, compared with full distilling operation, our TCCT architectures could help Informer deal with more complicated LSTF problems. (6) To strengthen the statement that dilated causal convolution and passthrough mechanism could strengthen the learning capability of Informer, we perform additional experiments under the setting of this experiment group where Informer with only dilated causal convolution(TCCT_IV), only passthrough mechanism(TCCT_V) or combined(TCCT_VI) are tested. The results are shown in Appendix C.

### 5.5. Forecasting capability in LSTI problem

Within this setting, long-sequence input processing capability of Informer, Informer + and TCCT_ III are evaluated. The former experiment has already shown the respective forecasting capability enhancement to Informer of three TCCT architectures so that TCCT_I and TCCT_ II do not need to take part in this experiment. This group of experiments are performed only using ETTm1 dataset under multivariate setting but with more input window sizes. We choose the prediction window size in {48, 96} while prolonging the encoder input length in {48, 96, 144, 192, 240, 288, 336, 384, 432} and {96, 192, 288, 384, 480, 576, 672, 768}.

From Table 3, it can be observed that: (1) TCCT_ III outperforms Informer and Informer + at nearly all cases showing that TCCT architectures still work when dealing with LSTI problems. (2) As the input length grows, the forecasting accuracy of Informer and Informer + decrease at some cases. Their maximum MSE drop are both over 10 %. Though TCCT_ III also has such cases, the number is smaller and maximum MSE drop is under 5 %. This phenomenon demonstrates that TCCT architectures could help Informer process longer sequence input and enhance Informer more robustness than full self-distilling operation.

#### 5.5.1. Ablation study on computation efficiency

We perform computation efficiency comparison to further explore differences of five models also only using ETTm1 dataset under multivariate setting. Training time and graphic memory usage are two criteria to measure the computation efficiency. We choose the prediction window size as 48 while prolonging the encoder input length in {48, 96, 144, 192, 240, 288, 336, 384, 432}. Note that in this group of experiments, training time of each model in each experiment is the summary of ten times rather than the average.

From Fig. 10(a), (b), it can be observed that: (1) Informer models with TCCT architectures use nearly the same training time and graphic memory at all cases, proving that the application of dilated causal convolution and passthrough mechanism would only bring negligible extra computation cost. (2) Informer models with TCCT architectures occupy less training time and memory usage than initial Informer and much less than Informer + . This phenomenon illustrates that TCCT architectures could improve the computation efficiency of Informer and are much more efficient than full distilling operation.

#### 5.5.2. Applicability analysis on other models

We perform applicability analysis of our proposed TCCT architectures on ETTh1 and ETTm1 under multivariate setting. Similar but more simplified compared to the first group of experiments, three TCCT architectures are simultaneously applied to canonical Transformer and LogTrans to inspect their applicability to enhancement of other Transformer models based on different self-attention mechanisms. All methods are using the baseline of Informer which consists of an encoder stacking three self-attention blocks and a 2-layer decoder. Methods related to canonical Transformer replace ProbSparse self-attention with canonical self-attention while methods related to LogTrans replace with LogSparse self-attention. Other hyper-parameters are similar to those of the first group of experiments.

From Table 4, we can observe that: (1) Both canonical Transformer and LogTrans behave better if combined with TCCT architectures. It is obvious that TCCT architectures are general and effective enough to cope with other Transformer models and self-attention mechanisms on time series forecasting. (2) TCCT architectures still keep their good qualities when coping with other Transformer models, including: a. Better forecasting performance with more complicated input; b. More robust forecasting accuracy. (3) Together with tables and figures above, it can be concluded that our TCCT architectures are effective, efficient, applicable and general enough to Transformer or Transformer-like models on time series forecasting.

#### 5.5.3. Applicability analysis on other datasets

In order to further evaluate the generalizability and applicability of our TCCT architectures, we perform experiments on {ETTh2, ETTm2} and two extra datasets in our last group of experiments. Two extra datasets are:

**Electricity consumption dataset**[3] **(ECL)**, after comprehensively assessing the dataset usage of Informer [11] and LogTrans [10], we eliminate the missing data in the dataset and convert the dataset into hourly electricity consumption (Kwh) of 321 clients of nearly 3 years. 'MT_321′ is set as the target value. The train/val/test is 21/7/7 months.

**Traffic usage dataset**[4] **(Traffic),** we choose traffic dataset but not weather dataset used in Informer experiments [11]. Because in Informer experiments performed on the weather dataset under univariate setting, the prediction results of different methods (Informer, LogTrans, LSTnet [15]) are quite close. Traffic usage dataset collects 24-month hourly data from the California Department of Transportation which describes the road occupancy rates (between 0 and 1) measured by different sensors on San Francisco Bay area freeways. 'Client_862' is set as the target value.

**Table 1**
Univariate long sequence time series forecasting results on ETTh1 and ETTm1.

| Methods | Informer | | Informer+ | | TCCT_I | | TCCT_ II | | TCCT_ III | |
|---|---|---|---|---|---|---|---|---|---|---|
| Prediction length | MSE | MAE | MSE | MAE | MSE | MAE | MSE | MAE | MSE | MAE |
| 48 | 0.1821 | 0.3611 | 0.1552 | 0.3220 | 0.1761 | 0.3453 | 0.1589 | 0.3261 | ***0.1245*** | ***0.2910*** |
| 96 | 0.2173 | 0.3952 | ***0.1811*** | 0.3635 | 0.2027 | 0.3805 | 0.1979 | 0.3738 | 0.1862 | ***0.3569*** |
| 192 | 0.2618 | 0.4309 | 0.2402 | 0.4183 | 0.2416 | 0.4165 | 0.2121 | 0.3895 | ***0.1995*** | ***0.3730*** |
| 384 | 0.2719 | 0.4513 | 0.2611 | 0.4499 | 0.2652 | 0.4367 | 0.2240 | 0.3935 | ***0.2154*** | ***0.3813*** |
| 48 | 0.1121 | 0.2819 | ***0.0603*** | ***0.1805*** | 0.1022 | 0.2712 | 0.0751 | 0.2378 | 0.0612 | 0.1849 |
| 96 | 0.1557 | 0.3381 | 0.1265 | 0.2951 | 0.1454 | 0.3108 | 0.1362 | 0.3080 | ***0.1245*** | ***0.2899*** |
| 192 | 0.2636 | 0.4324 | 0.2257 | 0.3961 | 0.2495 | 0.4151 | 0.2560 | 0.4122 | ***0.2186*** | ***0.3923*** |
| 384 | 0.3762 | 0.5590 | 0.3543 | ***0.5189*** | 0.3811 | 0.5396 | 0.3659 | 0.5430 | ***0.3502*** | 0.5216 |
| Counting | 0 | | 4 | | 0 | | 0 | | 12 | |

**Table 2**
Multivariate long sequence time series forecasting results on ETTh1 and ETTm1.

| Methods | Informer | | Informer+ | | TCCT_I | | TCCT_ II | | TCCT_ III | |
|---|---|---|---|---|---|---|---|---|---|---|
| Prediction length | MSE | MAE | MSE | MAE | MSE | MAE | MSE | MAE | MSE | MAE |
| 48 | 1.1309 | 0.8549 | 0.9483 | ***0.7157*** | 0.9737 | 0.7839 | 0.9694 | 0.7724 | ***0.8877*** | 0.7537 |
| 96 | 1.2433 | 0.9132 | 1.0575 | 0.8184 | 1.0761 | 0.8477 | 1.0578 | 0.8142 | ***1.0199*** | ***0.8069*** |
| 192 | 1.3011 | 0.9324 | 1.1477 | 0.8566 | 1.2101 | 0.8745 | 1.1785 | 0.8715 | ***1.1104*** | ***0.8458*** |
| 384 | 1.3313 | 0.9340 | 1.2665 | 0.8810 | 1.2284 | 0.8825 | 1.1913 | 0.8520 | ***1.1527*** | ***0.8356*** |
| 48 | 0.5282 | 0.5170 | 0.4890 | 0.4887 | 0.5172 | 0.4941 | 0.5036 | 0.4732 | ***0.4464*** | ***0.4354*** |
| 96 | 0.6596 | 0.5915 | 0.5867 | 0.5646 | 0.6101 | 0.5649 | 0.5811 | 0.5440 | ***0.5772*** | ***0.5424*** |
| 192 | 0.7687 | 0.6699 | 0.6683 | 0.5992 | 0.6854 | 0.6153 | 0.6510 | 0.5947 | ***0.6375*** | ***0.5823*** |
| 384 | 0.7996 | 0.6754 | 0.7650 | 0.6463 | 0.7812 | 0.6744 | 0.7460 | ***0.6222*** | ***0.7415*** | 0.6250 |
| Counting | 0 | | 1 | | 0 | | 1 | | 14 | |

**Table 3**
Multivariate long sequence time series input results on ETTm1 (Slash refers to out of memory).

| Methods | | Informer | | Informer+ | | TCCT_ III | |
|---|---|---|---|---|---|---|---|
| Prediction length | Input length | MSE | MAE | MSE | MAE | MSE | MAE |
| 48 | 48 | 0.6106 | 0.5440 | 0.5918 | 0.4994 | ***0.5527*** | ***0.4802*** |
| | 96 | 0.5449 | 0.4943 | 0.5685 | 0.4836 | ***0.4849*** | ***0.4637*** |
| | 144 | 0.5256 | 0.4718 | 0.5491 | 0.4853 | ***0.4754*** | ***0.4618*** |
| | 192 | 0.5089 | 0.4664 | 0.4996 | 0.4684 | ***0.4438*** | ***0.4428*** |
| | 240 | 0.4760 | 0.4617 | 0.5427 | 0.4849 | ***0.4335*** | ***0.4382*** |
| | 288 | 0.4698 | 0.4676 | 0.5318 | 0.4867 | ***0.4412*** | ***0.4370*** |
| | 336 | 0.5205 | 0.5550 | 0.4678 | 0.4592 | ***0.4515*** | ***0.4498*** |
| | 384 | 0.5525 | 0.5498 | ***0.4421*** | 0.4463 | 0.4448 | ***0.4386*** |
| | 432 | 0.4996 | 0.4923 | 0.5072 | 0.4992 | ***0.4419*** | ***0.4278*** |
| 96 | 96 | 0.6266 | 0.5829 | 0.6016 | 0.5511 | ***0.5900*** | ***0.5447*** |
| | 192 | 0.6052 | 0.5783 | 0.5863 | 0.5423 | ***0.5512*** | ***0.5273*** |
| | 288 | 0.5981 | 0.5680 | 0.5699 | 0.5354 | ***0.5415*** | ***0.5197*** |
| | 384 | 0.5686 | 0.5541 | 0.5724 | 0.5389 | ***0.5058*** | ***0.5082*** |
| | 480 | 0.5677 | 0.5459 | 0.5523 | 0.5358 | ***0.4960*** | ***0.5002*** |
| | 576 | 0.5965 | 0.5776 | 0.5601 | 0.5410 | ***0.4912*** | ***0.4995*** |
| | 672 | 0.5783 | 0.5549 | 0.5777 | 0.5215 | ***0.5150*** | ***0.5104*** |
| | 768 | 0.6439 | 0.5762 | | | ***0.5135*** | ***0.5029*** |
| Counting | | 0 | | 1 | | 33 | |

The train/val/test is 14/5/5 months. The split methods are the same as {ETTh1, ETTm1}, evenly and consecutively across time. Only Informer, Informer + and TCCT_ III are tested in this group of experiment. Other experiment settings are the same with first group of experiment, **Ablation Study on LSTF Problem**. More detailed information of datasets is shown in Appendix D.

From Tables 5 and 6, it can be observed that: (1) Even tested with other datasets, TCCT_ III still outperforms Informer and Informer + proving that TCCT architectures have good generalizability and applicability. (2) Similar characteristics of TCCT_ III are reflected compared with experiment results above, including relatively better performances when dealing with more complicated forecasting settings and robust forecasting accuracy. This phenomenon indicates that TCCT architectures have great possibility to stably enhance the forecasting accuracy of Transformer models not only on ETT datasets but also on other time series forecasting datasets. (3) It is proven again that full distilling operation is an unstable method due to the phenomenon that Informer + has worse performances compared with initial Informer in some occasions.

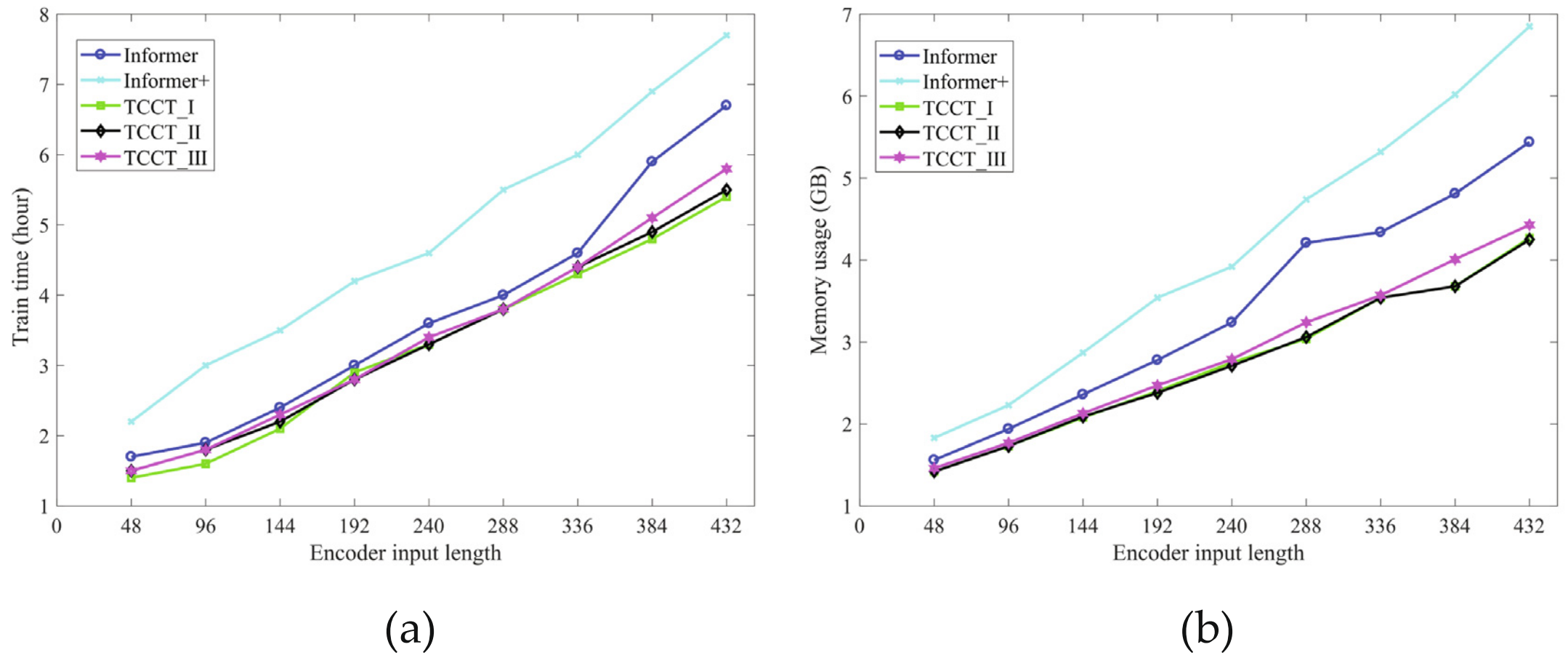

**Fig. 10.** Training time and memory usage results on ETTm1.

**Table 4**
Other methods' multivariate long sequence time-series forecasting results.

| Methods | Transformer | | Transformer + TCCT | | LogTrans | | LogTrans + TCCT | |
|---|---|---|---|---|---|---|---|---|
| Prediction length | MSE | MAE | MSE | MAE | MSE | MAE | MSE | MAE |
| 48 | 0.9377 | 0.7516 | ***0.7526*** | ***0.6426*** | 0.8303 | 0.7109 | ***0.7214*** | ***0.6424*** |
| 96 | 1.0683 | 0.8326 | ***0.9670*** | ***0.8131*** | 1.0499 | 0.8266 | ***1.0368*** | ***0.8248*** |
| 192 | 1.0706 | 0.8363 | ***1.0683*** | ***0.8218*** | 1.1883 | 0.8808 | ***1.1045*** | ***0.8488*** |
| 384 | 1.0977 | 0.8430 | ***1.0809*** | ***0.8286*** | ***1.1207*** | 0.8466 | 1.1210 | ***0.8414*** |
| 48 | 0.4811 | 0.4732 | ***0.4127*** | ***0.4404*** | 0.4598 | 0.4622 | ***0.4293*** | ***0.4563*** |
| 96 | 0.6256 | 0.6023 | ***0.5784*** | ***0.5684*** | 0.6061 | 0.5855 | ***0.5572*** | ***0.5478*** |
| 192 | 0.7219 | 0.6608 | ***0.6188*** | ***0.5837*** | 0.7219 | 0.6608 | ***0.7078*** | ***0.6450*** |
| 384 | 0.7849 | 0.6832 | ***0.7296*** | ***0.6536*** | 0.8271 | 0.7121 | ***0.7701*** | ***0.6732*** |
| Counting | 0 | | 16 | | 1 | | 15 | |

**Table 5**
Results on LSTF Problem under univariate setting and additional datasets.

| Methods | Informer | | Informer+ | | TCCT_ III | |
|---|---|---|---|---|---|---|
| Prediction length | MSE | MAE | MSE | MAE | MSE | MAE |
| 48 | 0.1964 | 0.3557 | ***0.1585*** | 0.3168 | 0.1596 | ***0.3113*** |
| 96 | 0.2732 | 0.4228 | 0.2359 | 0.3894 | ***0.2319*** | ***0.3717*** |
| 192 | 0.2873 | 0.4287 | 0.2608 | 0.4134 | ***0.2545*** | ***0.4121*** |
| 384 | 0.3085 | 0.4495 | 0.2828 | 0.4324 | ***0.2637*** | ***0.4204*** |
| 48 | 0.0769 | 0.2089 | 0.0667 | 0.1984 | ***0.0606*** | ***0.1885*** |
| 96 | 0.1279 | 0.2693 | 0.1037 | 0.2442 | ***0.0915*** | ***0.2310*** |
| 192 | 0.1692 | 0.3157 | 0.1497 | 0.3017 | ***0.1442*** | ***0.2981*** |
| 384 | 0.2163 | 0.3737 | ***0.2032*** | ***0.3549*** | 0.2080 | 0.3696 |
| 48 | 0.2644 | 0.3667 | ***0.2427*** | ***0.3637*** | 0.2482 | 0.3656 |
| 96 | 0.3016 | 0.3914 | 0.2866 | 0.3817 | ***0.2614*** | ***0.3714*** |
| 192 | 0.3440 | ***0.4096*** | 0.3489 | 0.4216 | ***0.3334*** | 0.4145 |
| 384 | 0.4150 | 0.4780 | ***0.3931*** | ***0.4582*** | 0.3950 | 0.4574 |
| 48 | 0.5257 | 0.5333 | 0.5876 | 0.5439 | ***0.5254*** | ***0.5295*** |
| 96 | 0.5791 | 0.5499 | 0.5475 | 0.5268 | ***0.5061*** | ***0.5047*** |
| 192 | 0.5581 | 0.5376 | 0.5364 | 0.5230 | ***0.5113*** | ***0.5143*** |
| 384 | 0.6376 | 0.5801 | 0.6459 | 0.5831 | ***0.6255*** | ***0.5424*** |
| Counting | 1 | | 7 | | 24 | |

## 6. Conclusion

In this paper, we proposed the concept of 'tight-coupled convolutional Transformer' (TCCT) and three TCCT architectures to improve the prediction capability of state-of-the-art Transformer models on time series forecasting. Specially, we designed CSPAttention to reduce the computation cost and memory usage of self-attention mechanism without losing forecasting accuracy. Also, the application of dilated causal convolution enables Transformer models to obtain exponentially receptive field growth. Last but not least, we employed passthrough mechanism to help Transformer models gain finer-grained information. The separate and extensive experiments on real-word datasets demonstrated that all three TCCT architectures could enhance the performance of Transformer models on time series forecasting in different ways.

**Table 6**
Results on LSTF Problem under multivariate setting and additional datasets.

| Methods | Informer | | Informer+ | | TCCT_ III | |
|---|---|---|---|---|---|---|
| Prediction length | MSE | MAE | MSE | MAE | MSE | MAE |
| 48 | 3.5159 | 1.6467 | ***3.1473*** | ***1.4802*** | 3.3223 | 1.5573 |
| 96 | 3.7512 | 1.7219 | ***3.4148*** | ***1.5950*** | 3.6329 | 1.6105 |
| 192 | 3.8553 | 1.6957 | 4.5768 | 1.8840 | ***3.7806*** | ***1.6690*** |
| 384 | 3.9481 | 1.7177 | 4.1071 | 1.7310 | ***3.8125*** | ***1.6938*** |
| 48 | 1.3967 | 0.9225 | ***0.5960*** | ***0.5955*** | 0.6435 | 0.6220 |
| 96 | 2.1326 | 1.1159 | 1.5289 | ***0.9206*** | ***1.4601*** | 0.9335 |
| 192 | 2.4674 | 1.2306 | 1.9355 | 1.0720 | ***1.8971*** | ***1.0305*** |
| 384 | 4.7351 | 1.7298 | 4.0677 | 1.6337 | ***3.7876*** | ***1.6095*** |
| 48 | 0.3285 | 0.3947 | 0.3282 | 0.3925 | ***0.3208*** | ***0.3895*** |
| 96 | 0.3532 | 0.4133 | 0.3490 | 0.4082 | ***0.3410*** | ***0.4028*** |
| 192 | 0.3803 | 0.4354 | ***0.3614*** | 0.4164 | 0.3635 | ***0.4145*** |
| 384 | 0.3996 | 0.4384 | 0.3866 | 0.4326 | ***0.3760*** | ***0.4262*** |
| 48 | 0.7939 | 0.4203 | 0.7496 | 0.4021 | ***0.7277*** | ***0.3837*** |
| 96 | 0.8182 | 0.4293 | 0.7850 | 0.4113 | ***0.7384*** | ***0.3870*** |
| 192 | 0.8685 | 0.4593 | 0.8519 | 0.4414 | ***0.7515*** | ***0.3996*** |
| 384 | 0.9711 | 0.5058 | 0.9278 | 0.4826 | ***0.7920*** | ***0.4283*** |
| Counting | 0 | | 8 | | 24 | |

## Funding

This research was funded by the National Key Research and Development Program of China [Grant No. 2020YFB0505602]; and the National Natural Science Foundation of China [grant number 62076019].

## CRediT authorship contribution statement

**Li Shen:** Conceptualization, Methodology, Software, Data curation, Writing – original draft, Visualization, Validation, Investigation, Writing – review & editing. **Yangzhu Wang:** Supervision, Funding acquisition, Resources.

## Declaration of Competing Interest

The authors declare that they have no known competing financial interests or personal relationships that could have appeared to influence the work reported in this paper.

## Appendix A. Proof of Theorem 1

Proof. Given two matrices $A \in R^{a\times b}$ and $B \in R^{b\times c}$, only considering the computation of the multiplication, time complexity of $A \times B$ is $a \times b \times c$, denoted by $T(A \times B) = a \times b \times c$. Let $X \in R^{L\times d}$ be a input matrix consisting of $L$ tokens and the self-attention block has $H$ heads. Excluding biases, the output of the $h^{\text{th}}$ self-attention head [39] can be written as

$$A_h(X) = P_h X W_{V,h}$$

$P_h$ is the scaled dot-product and is given by

$$P_h = softmax\left(d_{qk}^{-1/2}\left(XW_{Q,h}^T W_{K,h} X^T\right)\right)$$

In this way, ignoring the computation of softmax, the time complexity of a single self-attention head could be written as

$$T(A_h(X)) \approx T\left(XW_{Q,h}^T W_{K,h} X^T \times XW_{V,h}\right)$$

$$= L \times d \times \frac{d}{H} + L \times \frac{d}{H} \times d + L \times d \times L + L \times L \times d + L \times d \times \frac{d}{H}$$

$$= 3\frac{Ld^2}{H} + 2L^2 d$$

Considering the last linear projection layer and the concatenation of all heads, the whole time complexity of a canonical self-attention block is given by

$$T(SA(X)) = H \times T(A_h(X)) + Ld^2$$

$$= 4d^2L + 2HdL^2$$

However, when calculating the time complexity of CSPAttention, supposing that CSPAttention splits the input dimension in half, the first part of CSPAttention has only one linear projection layer, which means that the time complexity of the first part could be written as

$$T(CSPA_1(X_1)) = L \times (d/2)^2$$

And the second part's time complexity is given by

$$T(CSPA_2(X_2)) = 4L(d/2)^2 + 2L^2(d/2)$$

$$= Ld^2 + HL^2 d$$

So the total time complexity of CSPAttention is

$$T(CSPA(X)) = 1.25d^2L + HdL^2$$

It is obvious that the time complexity is approximately related to the coefficient of $L^2$ with the grow of $L$. So the time complexity of CSPAttention is almost $2Hd/Hd = 50\%$ of that of canonical self-attention when the network is propagating forward. Moreover, the coefficient of $L$ of CSPAttention's time complexity is $1.25d^2/4d^2 = 31.25\%$ of that of canonical self-attention's time complexity. Consequently, CSPAttention cuts down at least 50 % time complexity compared with canonical self-attention mechanism.

## Appendix B. Additional tables of MSE standard deviation

In tables and analysis below, MSD is short for MSE Standard Deviation and CV is short for coefficient of variation.

From Tables B1 and B2, it can be concluded that: (1) The results of all five methods' ten-turn experiments share similar consistency except several cases. They are relatively unstable under univariate setting, especially when prediction length is short. This is reasonable because average MSE losses under this occasion are also quite small, which makes turbulence more decisive. The phenomenon of better consistency under multivariate setting also proves that empirically relative time series in ETT dataset are truly relative and really help improve forecasting stability. (2) However, it can also be observed that the full distilling operation, which distinguishes Informer + and Informer with, makes model more unstable. The biggest MSE CV of informer + under univariate setting achieves exaggerate 48.42 %, which is much bigger than other methods' under the same setting. Under multivariate setting, Informer + is the only method has MSE CV over than 15 %. All these phenomena prove that Informer + is less stable than other methods and full distilling operation behaves worse than passthrough mechanism in the aspect of consistency. This point is identical to the analysis in the section of Passthrough Mechanism of Methodology.

**Table B1**
MSD and CV of the results in Table 1 (Univariate).

| Methods | Informer | | Informer+ | | TCCT_I | | TCCT_ II | | TCCT_ III | |
|---|---|---|---|---|---|---|---|---|---|---|
| Prediction length | MSD | CV (%) | MSD | CV (%) | MSD | CV (%) | MSD | CV (%) | MSD | CV (%) |
| 48 | 0.0173 | 9.50 | 0.0516 | 33.24 | 0.0184 | 10.45 | 0.0147 | 9.25 | 0.0153 | 12.29 |
| 96 | 0.0345 | 15.87 | 0.0224 | 12.36 | 0.0269 | 13.27 | 0.0244 | 12.33 | 0.0206 | 11.06 |
| 192 | 0.0382 | 14.59 | 0.0353 | 14.70 | 0.0356 | 14.74 | 0.0200 | 9.43 | 0.0235 | 11.78 |
| 384 | 0.0441 | 16.21 | 0.0219 | 8.38 | 0.0373 | 14.06 | 0.0258 | 11.52 | 0.0336 | 15.60 |
| 48 | 0.0383 | 34.12 | 0.0292 | 48.42 | 0.0248 | 24.27 | 0.0191 | 25.43 | 0.0154 | 25.16 |
| 96 | 0.0359 | 23.05 | 0.0214 | 16.92 | 0.0386 | 26.54 | 0.0421 | 30.91 | 0.0255 | 20.48 |
| 192 | 0.0422 | 16.01 | 0.0275 | 12.18 | 0.0301 | 12.06 | 0.0316 | 12.34 | 0.0208 | 9.51 |
| 384 | 0.0473 | 12.57 | 0.0318 | 8.97 | 0.0469 | 12.31 | 0.0368 | 10.06 | 0.0443 | 12.65 |

**Table B2**
MSD and CV of the results in Table 2 (Multivariate).

| Methods | Informer | | Informer+ | | TCCT_I | | TCCT_ II | | TCCT_ III | |
|---|---|---|---|---|---|---|---|---|---|---|
| Prediction length | MSD | CV (%) | MSD | CV (%) | MSD | CV (%) | MSD | CV (%) | MSD | CV (%) |
| 48 | 0.1465 | 12.95 | 0.1243 | 10.87 | 0.1031 | 10.59 | 0.0475 | 4.90 | 0.0366 | 4.12 |
| 96 | 0.0569 | 4.58 | 0.1870 | 17.68 | 0.0490 | 4.55 | 0.0468 | 4.42 | 0.0530 | 6.57 |
| 192 | 0.0464 | 3.57 | 0.1303 | 11.35 | 0.0649 | 5.36 | 0.0620 | 5.26 | 0.0741 | 6.67 |
| 384 | 0.0518 | 3.89 | 0.0601 | 4.75 | 0.0587 | 4.78 | 0.0409 | 3.43 | 0.0535 | 4.64 |
| 48 | 0.0124 | 2.35 | 0.0577 | 11.80 | 0.0149 | 2.88 | 0.0261 | 5.52 | 0.0358 | 8.02 |
| 96 | 0.0572 | 8.67 | 0.0382 | 6.51 | 0.0319 | 5.23 | 0.0396 | 6.81 | 0.0448 | 7.76 |
| 192 | 0.0498 | 6.48 | 0.0277 | 4.14 | 0.0363 | 5.30 | 0.0443 | 6.80 | 0.0222 | 3.48 |
| 384 | 0.0461 | 5.76 | 0.0264 | 3.45 | 0.0354 | 4.53 | 0.0408 | 6.56 | 0.0426 | 5.75 |

## Appendix C. Test of dilated causal convolution and passthrough mechanism

From Tables C1, C2, 1 and 2, we can observe that: (1) TCCT_VI, Informer coping with dilated causal convolution, outperforms Informer in most of occasions, indicating that dilated causal convolution really helps improve Informer's forecasting capability. However, hardly ever could TCCT_VI outperform Informer+, which is reasonable because TCCT_VI only uses one encoder and ignores former feature maps from encoder's previous self-attention blocks, which makes it have worse performance compared with Informer + or TCCT_V. (2) TCCT_V, Informer coping with passthrough mechanism, also outperforms Informer. Moreover, it outperforms TCCT_I and TCCT_VI at most of cases, indicating that passthrough mechanism could mostly enhance Informer's forecasting capability among all three TCCT architectures. (3) Though it seems that passthrough mechanism does better job than other two TCCT architectures, TCCT_V only has a little more than 50 % chance to outperform Infor-

**Table C1**
Results of additional methods on LSTF Problem under univariate setting.

| Methods | TCCT_VI | | TCCT_V | | TCCT_IV | |
|---|---|---|---|---|---|---|
| Prediction length | MSE | MAE | MSE | MAE | MSE | MAE |
| 48 | 0.1696 | 0.3411 | 0.1527 | 0.3180 | 0.1488 | 0.3190 |
| 96 | 0.1962 | 0.3684 | 0.1784 | 0.3534 | 0.1653 | 0.3322 |
| 192 | 0.2190 | 0.3988 | 0.2057 | 0.3777 | 0.1915 | 0.3746 |
| 384 | 0.2236 | 0.4077 | 0.2187 | 0.3983 | 0.2062 | 0.3810 |
| 48 | 0.1080 | 0.2661 | 0.1030 | 0.2531 | 0.0850 | 0.2162 |
| 96 | 0.1627 | 0.3335 | 0.1484 | 0.3227 | 0.1140 | 0.2776 |
| 192 | 0.2563 | 0.4165 | 0.2382 | 0.4037 | 0.2090 | 0.3769 |
| 384 | 0.4048 | 0.5572 | 0.3657 | 0.5476 | 0.3556 | 0.5050 |
| Win Informer | 14 | | 16 | | 16 | |
| Win Informer+ | 4 | | 8 | | 13 | |
| Win TCCT_ III | 0 | | 2 | | 8 | |

**Table C2**
Results of additional methods on LSTF Problem under multivariate setting.

| Methods | TCCT_VI | | TCCT_V | | TCCT_IV | |
|---|---|---|---|---|---|---|
| Prediction length | MSE | MAE | MSE | MAE | MSE | MAE |
| 48 | 1.0192 | 0.8041 | 0.9285 | 0.7291 | 0.9005 | 0.7601 |
| 96 | 1.0521 | 0.8145 | 1.0364 | 0.8065 | 1.0493 | 0.8026 |
| 192 | 1.1901 | 0.9036 | 1.1486 | 0.8658 | 1.1444 | 0.8319 |
| 384 | 1.2500 | 0.9140 | 1.1673 | 0.8707 | 1.1605 | 0.8627 |
| 48 | 0.4732 | 0.4843 | 0.4724 | 0.4617 | 0.4715 | 0.4665 |
| 96 | 0.6264 | 0.5921 | 0.5802 | 0.5576 | 0.5789 | 0.5562 |
| 192 | 0.6597 | 0.6140 | 0.6512 | 0.6142 | 0.6475 | 0.5938 |
| 384 | 0.7684 | 0.6564 | 0.7247 | 0.6260 | 0.7258 | 0.6155 |
| Win Informer | 15 | | 16 | | 16 | |
| Win Informer+ | 5 | | 12 | | 15 | |
| Win TCCT_ III | 0 | | 3 | | 3 | |

**Table D1**
More details of all datasets.

| | $ETTh_{1,2}$ | $ETTm_{1,2}$ | Electricity | Traffic |
|---|---|---|---|---|
| Sample Rate | hourly | every 15 min | hourly | hourly |
| Length | 17,420 | 69,680 | 26,304 | 17,545 |
| Number | 7 | 7 | 321 | 862 |

mer+. This phenomenon illustrates that passthrough mechanism may work better than full distilling operation but not much better. Only coping with more TCCT architectures could TCCT_V obviously outperform Informer+. It could be proved by the experiment results that TCCT_IV, Informer with dilated convolution and passthrough mechanism, and TCCT_ III, Informer with all three architectures, achieve better forecasting results.

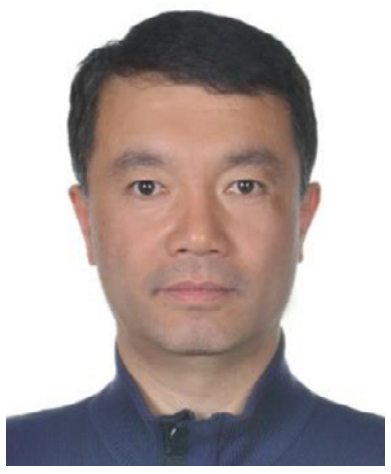

**Li Shen** received his B.S. degree in Navigation and Control from Beihang University in China at the School of Automation Science and Electrical Engineering. He is currently completing a MA.Eng degree in Electronic Information Engineering at the Beihang Institute of Unmanned System. His research interests include Computer Vision, Natural Language Processing and Time Series Forecasting.

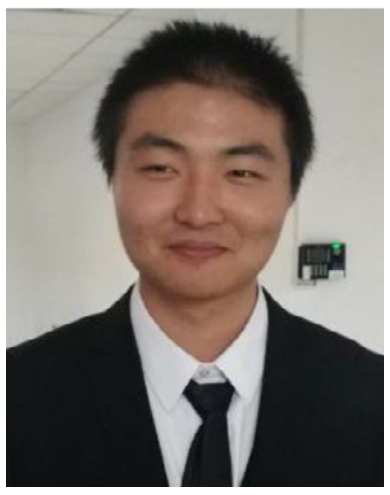

**Yangzhu Wang** received his B.S, MA.Eng and Ph.D degrees in Measurement and Control Technology and Instrumentation Program from Beihang University in China at the Beihang Institute of Unmanned System. He is currently a researcher fellow of the Flying College of Beihang University and Beihang Institute of Unmanned System. His research interests include Computer Vision, Unmanned System, and Measurement and Control Technology.